\documentclass[10pt,twocolumn,letterpaper]{article}

\usepackage{iccv}
\usepackage{times}
\usepackage{epsfig}
\usepackage{graphicx}
\usepackage{amsmath}
\usepackage{amssymb}
\usepackage{multirow}
\usepackage{bbding}
\usepackage{makecell}
\usepackage[accsupp]{axessibility}  % Improves PDF readability for those with disabilities.

\usepackage[pagebackref=true,breaklinks=true,letterpaper=true,colorlinks,bookmarks=false]{hyperref}

\iccvfinalcopy % *** Uncomment this line for the final submission

\def\iccvPaperID{10634} % *** Enter the ICCV Paper ID here

\ificcvfinal\fi

\begin{document}

%%%%%%%%% TITLE
\title{Predict to Detect: Prediction-guided 3D Object Detection using Sequential Images}
% Predict to Detect: Combining Prediction and Detection for Sequential Image-based 3D Object Detection
% Predict to Detect: Leveraging Predictions for 3D Object Detection using Sequential Images
% P2D: Leveraging Prediction and Detection for Motion-based 3D Object Detection
% P2D: Leveraging Prediction and Detection for Accurate Camera-based 3D Object Detection

\author{Sanmin Kim \hspace{15pt} Youngseok Kim \hspace{15pt} In-Jae Lee \hspace{15pt} Dongsuk Kum \\
KAIST\\
{\tt\small \{sanmin.kim, youngseok.kim, oliver0922, dskum\}@kaist.ac.kr} 
}

\maketitle
\ificcvfinal\thispagestyle{empty}\fi

%%%%%%%%% ABSTRACT
\begin{abstract}

Recent camera-based 3D object detection methods have introduced sequential frames to improve the detection performance hoping that multiple frames would mitigate the large depth estimation error.
Despite improved detection performance, prior works rely on naive fusion methods (e.g., concatenation) or are limited to static scenes (e.g., temporal stereo), neglecting the importance of the motion cue of objects.
These approaches do not fully exploit the potential of sequential images and show limited performance improvements.
To address this limitation, we propose a novel 3D object detection model, P2D (Predict to Detect), that integrates a prediction scheme into a detection framework to explicitly extract and leverage motion features.
P2D predicts object information in the current frame using solely past frames to learn temporal motion features. 
We then introduce a novel temporal feature aggregation method that attentively exploits Bird's-Eye-View (BEV) features based on predicted object information, resulting in accurate 3D object detection.
Experimental results demonstrate that P2D improves mAP and NDS by 3.0\% and 3.7\% compared to the sequential image-based baseline, proving that incorporating a prediction scheme can significantly improve detection accuracy.

\end{abstract}

%%%%%%%%% BODY TEXT
\section{Introduction}

\begin{figure*}[t]
\centering
\includegraphics[scale=1.07]{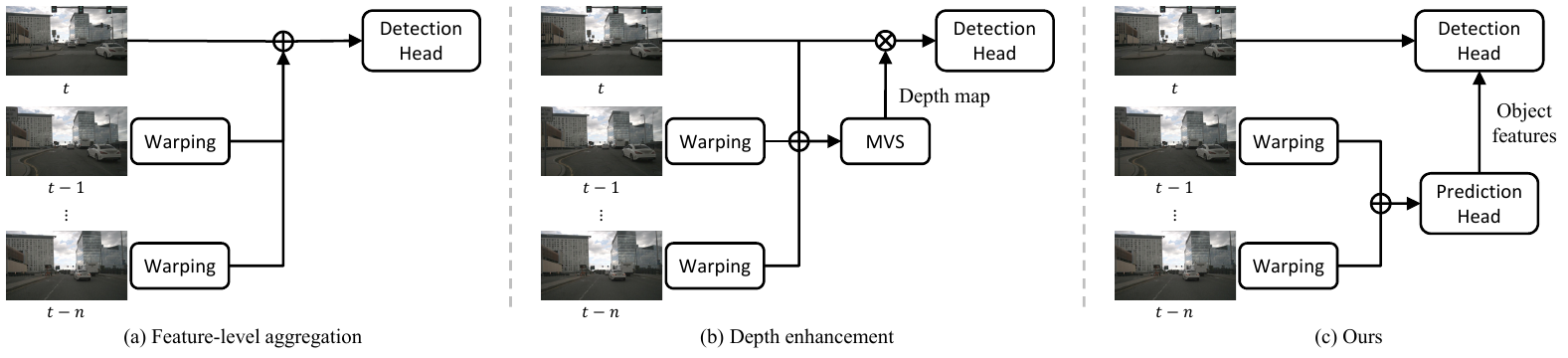}

\caption{
Comparison of temporal image-based methods. 
All methods align features by warping previous frames to the current frame.
(a) Feature-level aggregation methods naively concatenate sequential features before inputting them to the detection head. 
(b) Depth enhancement methods facilitate depth estimation using Multi-View Stereo (MVS). 
(c) Our proposed approach combines prediction and detection to leverage motion features. 
We predict object information from previous frames and use it to detect objects in the current frame.
}

\label{figure:1}
\vspace{-5pt}
\end{figure*}

3D object detection is an essential task for building a reliable self-driving system. 
In recent years, camera-based 3D object detection \cite{park2021dd3d, reading2021caddn,simonelli2019monodis, wang2021fcos3d} has gained widespread attention due to the cost-effectiveness of a camera sensor and its high-resolution characteristic. 
However, camera-based 3D object detection still has limited performance due to the scale ambiguity caused by projecting 3D space onto a 2D image and the absence of motion cues that are difficult to capture in a single image.
%===============================================================================================================================================================

Recent works have mitigated these drawbacks by leveraging multiple frames from history.
Multi-frame approaches incorporate temporal information into the space domain to provide richer information.
Moreover, sequence images are often readily available in real-world applications such as autonomous driving, making the use of sequence images an attractive option for performance improvements.

Previous works \cite{huang2022bevdet4d, jiang2023polarformer, li2023bevdepth, liu2022petrv2} have used temporal images in feature-level aggregation by concatenating sequential features to merge them. 
On the other hand, another line of work \cite{li2023bevstereo, park2022solofusion, wang2022dfm, wang2022sts} has adopted a temporal stereo \cite{yao2018mvsnet} to enhance depth estimation using the multi-view stereo (MVS) \cite{esteban2004silhouette}.
Although these methods have proved the effectiveness of sequence frames over a single frame, they did not thoroughly investigate into the motion cue of objects, from which the object detection would benefit by using sequence images.

Temporal images have rich motion information, which can provide critical motion features for accurate object detection. 
To further demonstrate the importance of motion cues in detection, we conducted experiments and evaluated the performance of the prediction-only results, which rely on motion prediction from previous frames, without using the current frame. 
Our findings from Table \ref{table:prediction_only} indicate that the prediction-only results (P) can achieve comparable performance to the final results (P+D), reaching up to 76\% and 89\% in terms of mAP and NDS, respectively. 
The final results (P+D) denote detection results that incorporate all temporal frames, including the current frame.
This experiment highlights the potential of using motion features from previous frames, which has been overlooked in prior works. 

To this end, we propose a novel sequential image-based 3D object detection model that learns motion cues to improve detection accuracy.
Our approach, P2D (Predict to Detect), introduces a prediction scheme into the detection task to fully exploit multi-frame image data. 
Specifically, P2D conducts motion prediction using previous frames to output the predicted objects' information for the current frame.
In the feature aggregation module, we employ a deformable attention \cite{zhu2020deformable} to make a spatio-temporal feature on the basis of prediction results that contains motion features.
Finally, the 3D detection head takes aggregated the spatio-temporal feature and outputs the final detection results.
In this way, our proposed method can fully benefit from multi-frame inputs by predicting objects' motion and utilizing it explicitly, providing a more accurate and reliable 3D object detection system for autonomous driving.

In summary, our contributions are as follows:
\begin{itemize}

\item We identify the motion feature as a key factor when handling sequential images for 3D object detection. A prediction mechanism is introduced to fully exploit the motion feature of multi-frame image data.
\vspace{-2pt}
\item We propose a novel 3D object detection model using sequential images. Our model includes a Prediction Head to predict object information and a Prediction-guided Feature Aggregation to integrate temporal features using motion features.
\vspace{-2pt}
\item Our approach achieves improved performance compared to prior state-of-the-art methods. Extensive experimentation confirms the effectiveness of our approach in adapting to moving objects and accurately estimating their velocities.

\end{itemize}

\section{Related Work}

\begin{figure*}[t]
\centering
\includegraphics[scale=1.04]{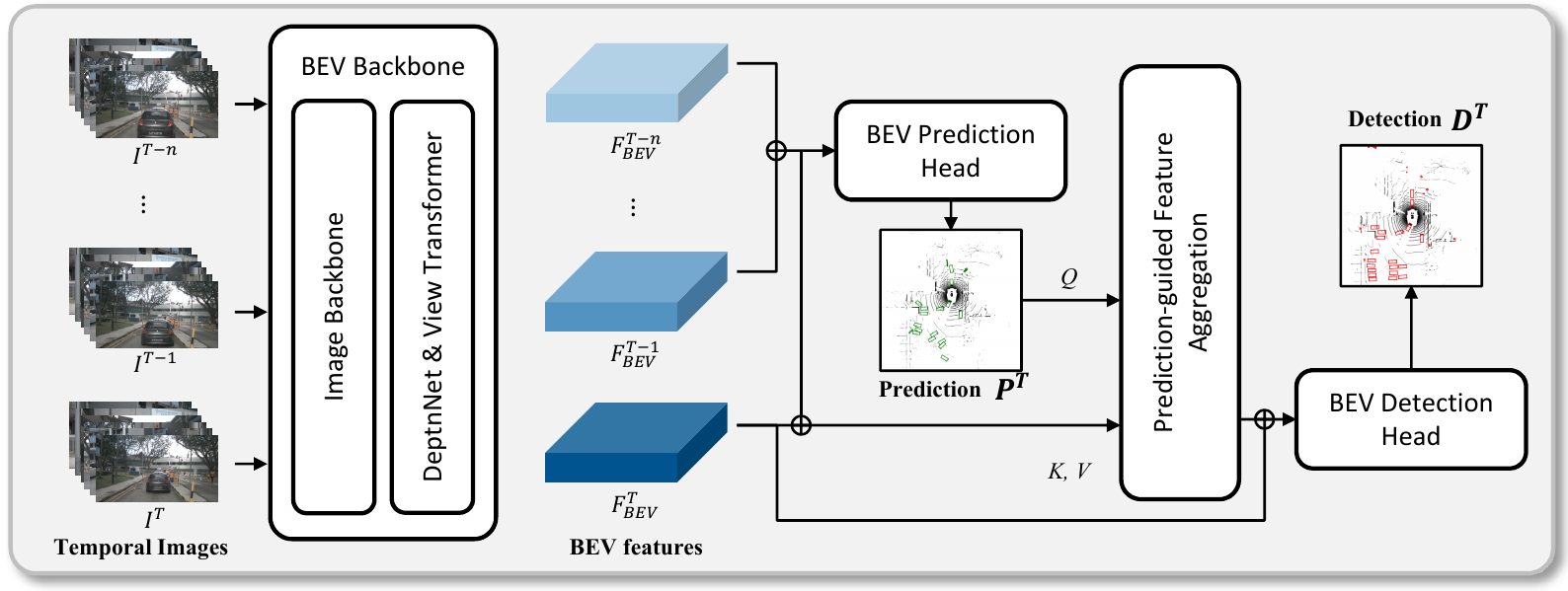}

\caption{
Overall architecture of P2D. The BEV backbone extracts BEV features from multi-view and multi-timestep images. The BEV Prediction Head takes BEV features of previous frames as the input and predicts objects in the current frame. The Prediction-guided Feature Aggregation module merges all temporal features based on predicted object information. The BEV Detection Head takes the aggregated feature and outputs the final detection results. The whole model is trained with the two loss terms of prediction and detection loss.
} 
\label{figure:architecture}
\vspace{-5pt}
\end{figure*}

%============================================================================================
\subsection{Camera-based 3D Object Detection}
Camera-based 3D object detection has gained significant attention following the success of 2D detection methods.
Early methods \cite{brazil2019m3drpn, duan2019centernet, liu2020smoke, mousavian20173d, park2021dd3d, wang2022pgd, wang2021fcos3d} exploit perspective view features by extracting 2D features from input images and directly estimating 3D information for object detection.
After the pioneering work of Mono3D \cite{chen2016mono3d}, M3D-RPN \cite{brazil2019m3drpn} proposes 3D anchor boxes and depth-aware convolution and FCOS3D \cite{wang2021fcos3d} projects 3D targets into a 2D image plane.
To mitigate the depth ambiguity of 2D images, several methods \cite{li2022monodde, lu2021gup, qin2022monoground, wang2022pgd} leverage geometric information while \cite{park2021dd3d} employs additional depth supervision.
PGD \cite{wang2022pgd} constructs geometric relation graphs across predicted objects to facilitate depth estimation and GUPNet \cite{lu2021gup} estimates a depth using height information.
On the other hand, DD3D \cite{park2021dd3d} boosts depth estimation ability using extra datasets \cite{guizilini2020ddad}.

Another stream of work employs view transformation to overcome the limitations of the perspective view. Several works transform image pixels into 3D point clouds using estimated depth information to take advantage of LiDAR detector \cite{wang2019pseudolidar, you2019pseudolidar++}. 
On the other hand, other works proposed to transform image features into voxel-like Bird's Eye View (BEV) features for 3D perception \cite{philion2020lss, reading2021caddn, roddick2018oft}.
BEVDet \cite{huang2021bevdet} uses LSS \cite{philion2020lss} based approach, which leverages the depth distribution to transform the perspective-view features into BEV space. BEVDepth \cite{li2023bevdepth} adds depth supervision using LiDAR point cloud, whereas BEVFormer \cite{li2022bevformer} adopts a deformable attention \cite{zhu2020deformable}.

%============================================================================================

\subsection{Sequential Image-based 3D Object Detection}
To improve 3D object detection performance, several works expanded the time horizon by leveraging temporally sequential frames. 
Sequential image-based 3D object detection can be categorized into object-centric and scene-centric methods.

\noindent \textbf{Object-centric methods.\hspace{0.2cm}} 
Inspired by the object tracking, these methods \cite{brazil2020kinematic3d, cheng2020motionloss, li2022time3d} employ object-level association to improve detection performance. 
Object-centric methods detect objects frame by frame and refine the detection results by matching objects.
Kinematic3D \cite{brazil2020kinematic3d} uses a 3D Kalman Filter to consider the kinematic motion of objects and update detection results. 
MotionLoss \cite{cheng2020motionloss} introduces patch-wise motion loss for temporal consistency. 
Time3D \cite{li2022time3d} adopts object-wise attention for temporal matching, where detections from the current frame operate as queries, and those from previous frames are keys and values.

\noindent \textbf{Scene-centric methods. \hspace{0.2cm}} 
These methods are subdivided into two again: feature-level aggregation and depth enhancement. 
Feature-level aggregation methods \cite{huang2022bevdet4d, jiang2023polarformer, li2023bevdepth, li2022bevformer} extract features from each image and aggregate temporal features before inputting them to the detection head.
\cite{huang2022bevdet4d, jiang2023polarformer, li2023bevdepth} aggregate sequential image features by concatenating them after temporal alignment using warping into the current timestep to compensate for the ego-motion. 
BEVFormer \cite{li2022bevformer} adopts temporal attention to aggregate temporal features based on BEV queries.
Depth enhancement methods \cite{li2023bevstereo, park2022solofusion, wang2022dfm, wang2022sts} employ temporal stereo, which extends Multi-view Stereo (MVS) \cite{bae2022magnet, yao2018mvsnet} into temporal images. 
By setting the translation of an ego agent as a baseline, two temporally nearby images have stereo correspondence that can be used in stereo matching.
DfM \cite{wang2022dfm} employs the temporal stereo in monocular 3D object detection with a theoretical analysis. 
BEVStereo \cite{li2023bevstereo} improves the temporal stereo with a sparse cost volume and an iterative algorithm inspired by MaGNet \cite{bae2022magnet}. 
STS \cite{wang2022sts} focuses on the multi-view cameras by allowing correspondence across cameras. 

While these methods have demonstrated improved performance compared to single-frame approaches, they have their own limitations. 
Object-centric methods heavily depend on frame-by-frame detection results, and thus, they are susceptible to propagating single-frame errors.
Feature-level aggregation methods cannot take full advantage of temporal features due to their naive aggregation methods (\textit{e.g.}, concatenation), and temporal stereo has limited performance on moving objects because of the static scene assumption.
Our proposed method overcomes these limitations by introducing prediction into the detection framework and explicitly leveraging motion cues.

\section{Method}

\subsection{Overall Architecture}
P2D extends BEVDepth \cite{li2023bevdepth} to perform prediction and detection within a single framework. 
As illustrated in Fig. \ref{figure:architecture}, our proposed P2D consists of a BEV backbone, prediction head, prediction-guided feature aggregation, and detection head. 
The BEV backbone extracts BEV features from the temporal input images. 
The BEV prediction head takes BEV features of previous frames as input and predicts object information in the current frame without relying on the current image. 
The predicted object information and BEV features are then combined into a spatio-temporal feature using the prediction-guided feature aggregation module. 
Finally, the BEV detection head generates the final detection results by utilizing both the current frame BEV feature and the spatio-temporal feature.

\subsection{BEV Backbone}
The BEV backbone of P2D consists of an image backbone, a depth network, and a view transformer. 
The input to the backbone is $N$ multi-view and $T$ multi-timestep images represented as $I=\{I^{t} \in \mathbb{R}^{N \times H \times W \times 3}, t=1,2, \ldots, T-1, T\}$.
First, the image backbone (\textit{e.g.}, ResNet \cite{he2016resnet} with FPN \cite{lin2017fpn}) extracts perspective-view features from the input images. 
Then, a depth network estimates per-image depth information from these features. 
Next, a view transformer lifts the perspective-view features into 3D space using the estimated depth and pools them to make BEV representations. 
The BEV features are represented as $F^{1:T}_{BEV} = \{F^{t}_{BEV} \in \mathbb{R}^{X_{f} \times Y_{f} \times C_{f}}, t=1,2, \ldots, T-1, T\}$, where $X_{f}$ and $Y_{f}$ denote the grid size and $C_{f}$ denotes the channel size of BEV features.

To alleviate the effect caused by ego-motion, we align the coordinates of BEV features from previous images into the current frame, following \cite{huang2022bevdet4d}. For more detailed information on our backbone, please refer to BEVDepth \cite{li2023bevdepth}.

\subsection{BEV Prediction Head}

Existing approaches that use temporal images \cite{huang2022bevdet4d, jiang2023polarformer, li2023bevdepth, liu2022petrv2} often concatenate all temporal features after alignment to compensate ego-motion.
Although simple and intuitive, such a naive strategy can not fully utilize temporal cues, limiting performance gain from temporal frames.
Even other approaches like temporal stereo \cite{li2023bevstereo, wang2022dfm, wang2022sts} handle temporal features more effectively by enhancing depth estimation with MVS, but they still overlook the importance of motion features. 
To fully benefit from previous frames, we introduce prediction into the detection framework.

The BEV prediction head uses BEV features only from previous frames to predict object information in the current frame, as follows:

\begin{equation}
\label{eq:BEV prediction head}
    P^{T} =  \Phi_{p} (F^{1:T-1}_{BEV}),
\end{equation}

\noindent 
where $P^{T} \in \mathbb{R}^{X_{f} \times Y_{f} \times C_{o}}$ represents predicted object information.
$C_{o}$ is the number of output attributes, including localization, dimension, velocity, orientation of an object, and per-class heatmaps. 
The per-class heatmaps show the probability of a specific class object in each position of the BEV feature. 
$\Phi_{p}$ is the detection network, such as the CenterPoint \cite{yin2021centerpoint} head.

The prediction result $P^T$ provides valuable object-level information, including motion cues, to the downstream network such as the feature aggregation and detection head. It allows the model to leverage explicit object-level motion features. 
Moreover, supervision on the features from previous frames using the ground truth objects in the current frame can help the model learn the beneficial features of previous frames for detecting the current frames when aggregated. 
The impact of prediction supervision on the BEV backbone is reported in Table \ref{table:supervision}.

\begin{figure}[t]
\centering
\includegraphics[width=0.475\textwidth]{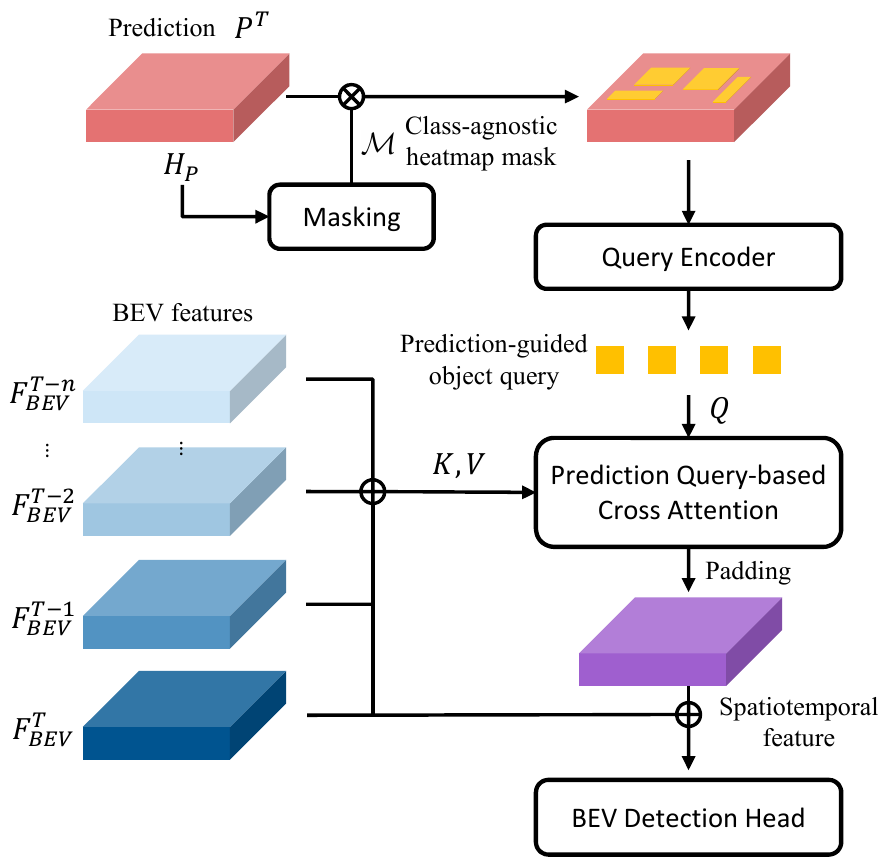}
\caption{
Illustration of the proposed Prediction-guided Feature Aggregation. The class-agnostic heatmap mask is generated from the class heatmap in the prediction results. The query encoder takes the masked prediction results to make the prediction-guided object queries. The prediction query-based cross-attention is a deformable attention module that uses object queries with keys and values from BEV features and fuses the temporal features.
} 
\label{figure:PFA}
\vspace{-5pt}
\end{figure}

\subsection{Prediction-guided Feature Aggregation}
Feature aggregation is a crucial module for effectively merging predicted object information and BEV features. 
To aggregate BEV features based on predicted object information, we introduce Prediction-guided object queries and Prediction query-based cross attention, as shown in Fig. \ref{figure:PFA}.

\noindent \textbf{Prediction-guided object queries. \hspace{0.2cm}}
In contrast to BEV queries in BEVFormer \cite{li2022bevformer}, which only has positional information on the BEV space at the initialization stage, we use prediction results $P^{T}$ as queries to gather temporal features based on predicted object information.
However, $P^{T}$ has a large dimension of $\mathbb{R}^{X_{f} \times Y_{f} \times C_{o}}$, which covers all locations of the BEV space, while objects only occupy a small region of it. 
Therefore, using $P^{T}$ as queries in its original form is highly inefficient in terms of computational cost.

\begin{figure}[t]
\centering
\includegraphics[scale=0.98]{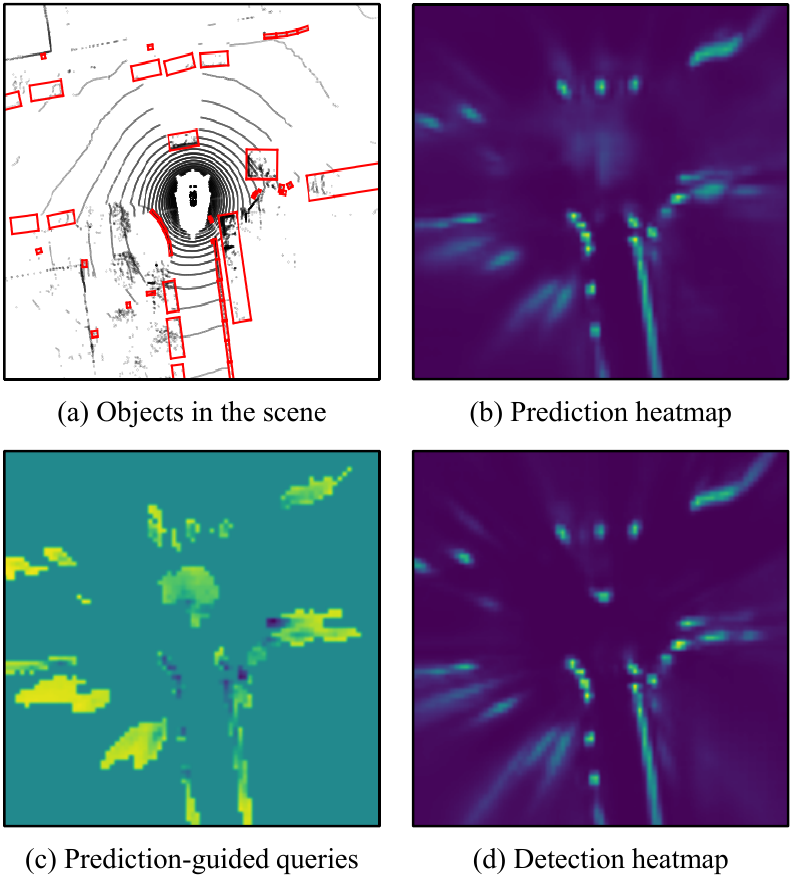}

\caption{
Visualization of the class-agnostic heatmap and prediction-guided object queries. (a) A sample frame with objects in the scenes. (b) A class-agnostic heatmap $H_p$, which is the output of the prediction head. (c) Prediction-guided object queries generated using the query mask and the predicted object information. (d) The heat map from the final detection results.
} 
\label{figure:heatmap}
\end{figure}

To address this problem, we use the object heatmap to select queries. The heatmap represent the probability that objects can exist in a specific space
Specifically, we extract a per-class heatmap represented as $H_{P} \in \mathbb{R}^{X_{f} \times Y_{f} \times N_{c}}$ from $P^{T}$ and generate the class-agnostic heatmap represented as $H_{CA} \in \mathbb{R}^{X_{f} \times Y_{f}}$ by selecting the maximum probability across all classes as follow:

\begin{equation}
\label{eq:class_ag heatmap}
    H_{CA}(i,j) = \max_{N_c}H_{P}(i,j),
\end{equation}

\noindent
where $(i, j)$ is the heatmap location index and $N_c$ is the number of object classes.

We then create a class-agnostic query mask $\mathcal{M} \in \mathbb{R}^{X_{f} \times Y_{f}}$ by filtering heatmap values with a threshold $\tau_{k}$.

\begin{equation}
\label{eq:query mask}
    \mathcal{M}(i,j) = \begin{cases}
    1 & H_{CA}(i,j) \ge \tau_{k}, \\
    0 & H_{CA}(i,j) < \tau_{k}.
    \end{cases}
\end{equation}

$\tau_{k}$ in Eq. \ref{eq:query mask} stands for an adaptive threshold for object probabilities.
We choose $\tau_{k}$ as the minimum value of top-k probabilities among $H_{CA}$ so that the number of queries can be fixed. 
The binary mask indicates the location candidates likely to be occupied by objects.

We apply the query mask to the prediction results to filter out less likely locations.
Finally, we embed the masked prediction results into queries using a linear projection.

\begin{equation}
\label{eq:queries}
    Q = \Phi_{q}(\mathcal{M} \odot P^{T}),
\end{equation}

\noindent
where $Q \in \mathbb{R}^{K \times C_{q}}$ stands for the selected prediction-guided object queries, $\Phi_{q}$ is a linear project for query embedding, $\odot$ operation denotes element-wise multiplication and $K$ is the number of queries.
In this way, we can avoid querying from empty space and reduce computational costs ($K  \ll X_{f}Y_{f}$).
We visualize a sample of the class-agnostic heatmap from both prediction and detection, and masked prediction-guided queries, in Fig. \ref{figure:heatmap}.

%==========================================================================================================================================
% nuscenes val set
%==========================================================================================================================================

\begin{table*}[t]

\begin{center}
\resizebox{0.98\textwidth}{!}{
\setlength{\tabcolsep}{5pt}
\renewcommand{\arraystretch}{1.3}
% \resizebox{\textwidth}{!}

\begin{tabular}{l|ccc|cc|ccccc}

\Xhline{2\arrayrulewidth}

\textbf{Method} & \textbf{Temporal} & \textbf{Backbone} & \textbf{Image Size} & \textbf{mAP} $\uparrow$ & \textbf{NDS} $\uparrow$& \textbf{mATE} $\downarrow$& \textbf{mASE} $\downarrow$& \textbf{mAOE} $\downarrow$& \textbf{mAVE} $\downarrow$& \textbf{mAAE} $\downarrow$\\ 
\hline
\hline

PETR$^{\dagger} \cite{liu2022petr}$  &  & ResNet50 & 384 $\times$ 1056   & 0.313 & 0.381 & 0.768 & 0.278 & 0.564 & 0.923 & 0.225 \\
BEVDet$^{\dagger}$ \cite{huang2021bevdet}&  & ResNet50 & 256 $\times$ 704    & 0.298 & 0.379 & 0.725 & 0.279 & 0.589 & 0.860 & 0.245 \\
BEVDet4D \cite{huang2022bevdet4d}        & \checkmark  & ResNet50 & 256 $\times$ 704    & 0.323 & 0.453 & 0.674 & 0.272 & \textbf{0.503} & 0.429 & \textbf{0.208} \\
BEVDepth \cite{li2023bevdepth}       & \checkmark  & ResNet50 & 256 $\times$ 704    & 0.333 & 0.441 & 0.683 & 0.276 & 0.545 & 0.526 & 0.226 \\
BEVStereo \cite{li2023bevstereo}     & \checkmark  & ResNet50 & 256 $\times$ 704    & 0.344 & 0.449 & 0.659 & 0.276 & 0.579 & 0.503 & 0.216 \\ 
\hline
P2D (BEVDepth)  & \checkmark  & ResNet50 & 256 $\times$ 704    & 0.360 & 0.474 & 0.643 & \textbf{0.271} & 0.512 & 0.412 & 0.217 \\ 
P2D (BEVStereo) & \checkmark  & ResNet50 & 256 $\times$ 704    & \textbf{0.374} & \textbf{0.486} & \textbf{0.631} & 0.272 & 0.508 & \textbf{0.384} & 0.212 \\ 
\hline
FCOS3D \cite{wang2021fcos3d}        &  & ResNet101 & 900 $\times$ 1600    & 0.295 & 0.372 & 0.806 & 0.268 & 0.511 & 1.131 & 0.170 \\
DETR3D$^\dagger$  \cite{wang2022detr3d}       &  & ResNet101 & 900 $\times$ 1600    & 0.349 & 0.434 & 0.716 & 0.268 & 0.379 & 0.842 & \textbf{0.200} \\
PETR$^\dagger$  \cite{liu2022petr}       &  & ResNet101 & 512 $\times$ 1408    & 0.357 & 0.421 & 0.710 & 0.270 & 0.470 & 0.885 & 0.224 \\
UVTR$^\dagger$  \cite{li2022uvtr}       & \checkmark & ResNet101 & 900 $\times$ 1600    & 0.379 & 0.483 & 0.731 & 0.267 & \textbf{0.350} & 0.510 & 0.200 \\
PolarDETR-T \cite{chen2022polardetr}    & \checkmark  & ResNet101 & 900 $\times$ 1600    & 0.383 & 0.488 & 0.707 & 0.269 & 0.344 & 0.518 & 0.196 \\
BEVDepth$^*$ \cite{li2023bevdepth}       & \checkmark  & ResNet101 & 512 $\times$ 1408    & 0.406 & 0.490 & 0.626 & 0.278 & 0.513 & 0.489 & 0.226 \\
BEVDStereo$^*$ \cite{li2023bevstereo}       & \checkmark  & ResNet101 & 512 $\times$ 1408    & 0.409 & 0.494 & 0.651 & 0.277 & 0.481 & 0.451 & 0.215 \\
BEVFormer \cite{li2022bevformer}     &\checkmark  & ResNet101 & 900 $\times$ 1600    & 0.416 & 0.517 & 0.673 & 0.274 & 0.372 & 0.394 & 0.198 \\
\hline
P2D (BEVDepth) & \checkmark  & ResNet101 & 512 $\times$ 1408    & 0.420 & 0.514 & \textbf{0.608} & 0.268 & 0.447 & 0.431 & 0.212 \\
P2D (BEVStereo) & \checkmark  & ResNet101 & 512 $\times$ 1408    & \textbf{0.433} & \textbf{0.528} & 0.619 & \textbf{0.265} & 0.432 & \textbf{0.364} & 0.211 \\
\hline
BEVDepth$^*$ & \checkmark  & ConvNext-B & 640 $\times$ 1600    & 0.426 & 0.521 & 0.587 & 0.267 & 0.393 & 0.444 & 0.229 \\
P2D (BEVDepth) & \checkmark  & ConvNext-B & 640 $\times$ 1600    & \textbf{0.460} & \textbf{0.551} & \textbf{0.537} & \textbf{0.259} & 0.398 & \textbf{0.388} & \textbf{0.212} \\
\Xhline{2\arrayrulewidth}

\end{tabular}
}
\end{center}
\vspace{-5pt}
\caption{Comparison on the nuScenes \textit{val} set. $\dagger$: methods with CBGS \cite{zhu2019cbgs}. $^*$: We reproduce the model without CBGS for a fair comparison.} 
\label{table:valset}

\end{table*}
%==========================================================================================================================================

\noindent \textbf{Prediction query-based cross attention. \hspace{0.2cm}}
Temporal features of a moving object are not projected into the same location even after alignment.
To effectively aggregate these temporal features, we adopt a deformable attention \cite{zhu2020deformable} by setting prediction-guided object queries $Q$ as queries and BEV features $F^{1:T}_{BEV}$ as keys and values.
We model the cross attention for temporal BEV features as follows:

\vspace{-7pt}

\begin{equation}
\label{eq:pqca}
\begin{split}
    & PQCA(Q_p, \{F^{1:T}_{BEV}\}) \\
    & = \sum_{t=1}^{T} DeformAttn(Q_{p}, p, F^{t}_{BEV}+e),
\end{split}
\end{equation}

\noindent where $Q_{p}$ denote a query located at $p=(i,j)$, respectively. $e=e_{s}+e^{t}$ denotes an embedding for positional and temporal dimensions. $t$ is the temporal indexes.
Additionally, we utilize zero-padding to match the shape of the output of PQCA with BEV feature's.

We stack each temporal feature level-wise and apply deformable cross attention across all timsteps. 
Through this, we can make a spatio-temporal feature by collecting related features in both spatial and temporal dimensions.
Our Prediction-guided Feature Aggregation is more effective in modeling the spatio-temporal feature compared to other aggregation methods such as stacking BEV features \cite{huang2022bevdet4d, li2023bevdepth} or Temporal Self-Attention \cite{li2022bevformer}.
This is because the prediction-guided object queries provide an object-level prior to the model and work as motion feature-based anchors for the cross-attention mechanism.

The BEV Detection Head concatenates the output of the Prediction Query-based Cross Attention with the BEV feature at the current frame.

\vspace{-7pt}

\begin{equation}
\label{eq:BEV detection head}
    D^{T} =  \Phi_{d} (PQCA(Q, \{F^{1:T}_{BEV}\}) \oplus F^{T}_{BEV}),
\end{equation}

\noindent where $D^{T}$ is the final detection output and $\Phi_{d}$ is the detection network  (\textit{e.g.}, CenterPoint \cite{yin2021centerpoint} head). 
It has the same network structure and outputs as BEV Prediction Head but does not share weights.

\subsection{Training}
P2D is an end-to-end trainable network, and the loss includes two terms: detection loss and prediction loss.

\vspace{-3pt}
\begin{equation}
\label{eq:loss1}
    \mathcal{L}= \mathcal{L}_{det} + \lambda_{p}\mathcal{L}_{pred},
\end{equation}

\noindent where $\lambda_{p}$ is balancing weight term. 
The detection loss $\mathcal{L}_{det}$ consists of classification loss, bounding box loss, and depth estimation loss. Meanwhile, the prediction loss $\mathcal{L}_{pred}$ contains two of them, except for depth estimation loss. 
We adopt the loss functions as focal loss for classification, L1 loss for bounding box regression, and binary cross-entropy for depth estimation.
It is worth noting that P2D does not require any additional annotations.

\section{Experiment}

%==========================================================================================================================================
% nuscenes testset results
%==========================================================================================================================================
\begin{table}[t]
\begin{center}
\resizebox{0.47\textwidth}{!}
{
\setlength{\tabcolsep}{2.5pt}
\renewcommand{\arraystretch}{1.3}
\begin{tabular}{l|c|cc|cccc}
\Xhline{2\arrayrulewidth}

\textbf{Method} & \textbf{Backbone} & \textbf{mAP} $\uparrow$& \textbf{NDS} $\uparrow$& \textbf{mATE} $\downarrow$& \textbf{mAOE}$\downarrow$\\ 
\hline
\hline
BEVDepth$^*$ & ResNet101 & 0.396 & 0.483 & 0.593 & 0.533 \\
BEVStereo$^*$ & ResNet101 & 0.404 & 0.502 & 0.587 & 0.518 \\
\hline
P2D(BEVDepth) & ResNet101 & 0.425 & 0.516 & \textbf{0.549} & 0.520 \\
P2D(BEVStereo) & ResNet101 & \textbf{0.436} & \textbf{0.530} & 0.550 & \textbf{0.517} \\

\Xhline{2\arrayrulewidth}
\end{tabular}
}
\end{center}
\vspace{-5pt}
\caption{Comparison on the nuScenes \textit{test} set. $^*$: We reproduce the model without CBGS for a fair comparison. } 

\label{table:testset}
\end{table}
%==========================================================================================================================================

\subsection{Dataset and Metrics}
We conduct experiments on the nuScenes dataset \cite{caesar2020nuscenes}, which consists of 1000 videos of around 20 seconds with annotations of 2Hz. 
The videos are split into three: 700, 150, and 150 scenes for training, validation, and testing, respectively. 
For the detection task, annotations contain 1.4M 3D bounding boxes of 10 object classes. We adopt the official evaluation metrics to evaluate performance, including nuScenes Detection Score (NDS), mean Average Precision (mAP), mean Average Translation Error (mATE), mean Average Scale Error (mASE), mean Average Orientation Error (mAOE), mean Average Velocity Error (mAVE), and mean Average Attribute Error (mAAE).

\subsection{Implementation Details}
Unless otherwise specified, we adopt BEVDepth \cite{li2023bevdepth} as our baseline model with the ImageNet pretrained ResNet50 \cite{he2016resnet} backbone and the input image size is resized to 256 $\times$ 704. We set the default BEV grid size as 128 $\times$ 128.
We follow image and BEV data augmentation strategies in \cite{li2023bevdepth}.

We use two previous frames with 1 second of time interval and set the number of object queries $k$ as 2048. We balance the loss function by setting $\lambda_{p}$ as 0.5. 
We trained the model using AdamW optimizer \cite{loshchilov2017adamw} for 24 epochs with a batch size of 16 on 4 NVIDIA 3090Ti GPUs. The learning rate is set to 2e-4, and the EMA technique is also used.

\subsection{Main Results}
We compare our model with existing camera-based detection models on the nuScenes validation dataset \cite{caesar2020nuscenes}. 
We report the results of P2D with two different baselines: BEVDepth \cite{li2023bevdepth} and BEVStereo \cite{li2023bevstereo} in Table \ref{table:valset}. 
The evaluation results demonstrate that P2D outperforms other methods and baselines significantly with ResNet50 backbone. 
Specifically, P2D achieves 2.7\% and 3.0\% improvement in mAP and 3.3 and 3.7 points improvement in NDS over BEVDepth and BEVStereo, respectively, outperforming other methods with a large margin.
In addition, P2D brings a substantial performance boost on velocity estimation, improving by 0.114 m/s and 0.119 m/s (21.5\% and 23.7\%) in mAVE compared to each baseline.  
In the case of using a larger backbone and input image size (ResNet101 \cite{he2016resnet} with 512 $\times$ 1408 and ConvNext-B \cite{liu2022convnet} with 640 $\times$ 1600), P2D consistently outperforms baselines and other methods both in mAP and NDS.

%==========================================================================================================================================
% prediction-only results
%==========================================================================================================================================
\begin{table}[t]
\begin{center}
\resizebox{0.47\textwidth}{!}
{
\setlength{\tabcolsep}{2pt}
\renewcommand{\arraystretch}{1.3}
\begin{tabular}{l|c|cc|cccc}
\Xhline{2\arrayrulewidth}

\textbf{Baseline} & \textbf{Strategy} & \textbf{mAP} $\uparrow$& \textbf{NDS} $\uparrow$& \textbf{mATE} $\downarrow$& \textbf{mAOE}$\downarrow$ & \textbf{mAVE} $\downarrow$\\ 
\hline
\hline
\multirow{2}{*}{BEVDepth} & P + D & 0.360 & 0.474 & 0.643 & 0.512 & 0.412 \\
& P & 0.272 & 0.422 & 0.740 & 0.587 & 0.294 \\
\hline
\multirow{2}{*}{BEVStereo} & P + D & 0.374 & 0.486 & 0.631 & 0.508 & 0.384 \\
& P & 0.272 & 0.418 & 0.738 & 0.606 & 0.308 \\

\Xhline{2\arrayrulewidth}

\end{tabular}
}
\end{center}
\vspace{-5pt}
\caption{Evaluation of prediction-only results on the nuScenes \textit{val} set. P+D represents the model with both prediction and detection, which is the same as our proposed P2D model. P represents the prediction-only results generated by the BEV Prediction Head, which does not use the current frame. 
} 
\label{table:prediction_only}
\end{table}
%==========================================================================================================================================

As shown in Table \ref{table:testset}, we also compare the performance with nuScenes test dataset.
With the same backbone and image size (ResNet101 with 512 $\times$ 1408), P2D still shows improved performance of 2.9\% and 3.2\% in mAP and 3.3 and 2.8 points in NDS over BEVDepth and BEVStereo, respectively, proving the effectiveness of the proposed method.

%==========================================================================================================================================
%% moving objects (vel > 1.0m/s)
%==========================================================================================================================================

\begin{table}[t]

\begin{center}
\resizebox{0.47\textwidth}{!}
{
\setlength{\tabcolsep}{3.5pt}
\renewcommand{\arraystretch}{1.3}
\begin{tabular}{l|ccccc}
\Xhline{2\arrayrulewidth}

\textbf{Methods} & \textbf{mATE} $\downarrow$& \textbf{mASE} $\downarrow$ & \textbf{mAOE} $\downarrow$& \textbf{mAVE} $\downarrow$& \textbf{mAAE} $\downarrow$\\ 
\hline
\hline
BEVDepth    &  0.815 & 0.271 & 0.404 & 2.010 & 0.159 \\
+ P2D       &  \textbf{0.783} & \textbf{0.256} & \textbf{0.367} & \textbf{1.712} & \textbf{0.149} \\
\hline
BEVStereo   &  0.822 & 0.269 & 0.345 & 1.712 & 0.149 \\
+ P2D       &  \textbf{0.773} & \textbf{0.260} & \textbf{0.243} & \textbf{1.477} & \textbf{0.146} \\

\Xhline{2\arrayrulewidth}
\end{tabular}
}
\end{center}
\vspace{-5pt}
\caption{Results on the moving objects. Only objects with a velocity higher than 1m/s are evaluated.} 
\label{table:moving_object}
\end{table}
%==========================================================================================================================================

%==========================================================================================================================================
%Ablation
%==========================================================================================================================================

\begin{table}[t]
\begin{center}
\resizebox{0.43\textwidth}{!}
{
\setlength{\tabcolsep}{4pt}
\renewcommand{\arraystretch}{1.3}
\begin{tabular}{cc|cc|ccc}
\Xhline{2\arrayrulewidth}

\textbf{PH} & \textbf{PFA} & \textbf{mAP} $\uparrow$&\textbf{NDS} $\uparrow$& \textbf{mATE} $\downarrow$& \textbf{mAOE} $\downarrow$& \textbf{mAVE} $\downarrow$\\ 
\hline
\hline
            &             & 0.334 & 0.448 & 0.680 & 0.551 & 0.469\\
 \checkmark &             & 0.351 & 0.466 & 0.668 & 0.528 & 0.414\\
            & \checkmark  & 0.353 & 0.459 & 0.662 & 0.541 & 0.472\\
 \checkmark & \checkmark  & \textbf{0.360} & \textbf{0.474} & \textbf{0.643} & \textbf{0.512} & \textbf{0.412}\\

\Xhline{2\arrayrulewidth}
\end{tabular}
}
\end{center}
\vspace{-5pt}

\caption{Ablation study on P2D. PH and PFA denote Prediction Head and Prediction-guided Feature Aggregation, respectively.} 
\label{table:ablation}

\end{table}
%==========================================================================================================================================

%==========================================================================================================================================
%number of input frames
%==========================================================================================================================================

\begin{table}[t]
\begin{center}
\setlength{\tabcolsep}{4pt}
\resizebox{0.47\textwidth}{!}
{
\renewcommand{\arraystretch}{1.3}
\begin{tabular}{l|c|cc|ccc}
\Xhline{2\arrayrulewidth}

\textbf{Methods} & \begin{tabular}[c]{@{}l@{}}\textbf{Prev.} \\ \textbf{frames}\end{tabular} & \textbf{mAP} $\uparrow$  &\textbf{ }\textbf{NDS} $\uparrow$  & \textbf{mATE} $\downarrow$ & \textbf{mAOE} $\downarrow$ & \textbf{mAVE} $\downarrow$ \\
\hline
\hline
\multirow{4}{*}{BEVDepth} & 0 & 0.312 & 0.357 & 0.695 & 0.645 & 1.144 \\
& 1 & 0.333 & 0.441 & 0.683 & 0.545 & 0.526 \\
& 2 & 0.334 & 0.448 & 0.680 & 0.551 & 0.469 \\
& 3 & 0.346 & 0.451 & 0.687 & 0.575 & 0.461 \\
\hline
\multirow{2}{*}{P2D} & 2 & 0.351 & 0.457 & 0.672 & 0.558 & \textbf{0.436}\\
& 3 & \textbf{0.362} & \textbf{0.465} & \textbf{0.652} & \textbf{0.434} & 0.464 \\

\Xhline{2\arrayrulewidth}

\end{tabular}
}
\end{center}
\vspace{-5pt}

\caption{Experiments on a different number of previous images.} 
\label{table:previ_frames}

\end{table}
%==========================================================================================================================================

\noindent \textbf{Prediction ability. \hspace{0.2cm}}
The quality of the prediction results generated from the prediction head plays a crucial role since it represents the potential of motion features.
In addition, it is even more important because P2D uses these prediction results as object queries.
Therefore, We evaluated the prediction-only results $P^{T}$ (the output of the prediction head) and reported them in Table \ref{table:prediction_only} to verify the effectiveness of the prediction.  
The results show that the prediction-only results can achieve comparable performance to the final detection results, up to 76\% and 89\% in mAP and NDS, respectively. 
It proves that the previous frames can estimate objects in the current frame using their motion features, and thus, the motion features can help the model to improve its detection performance.

\noindent \textbf{Moving objects. \hspace{0.2cm}}
In the autonomous driving environment, moving objects should be handled more attentively than static objects because moving objects often interact with autonomous agents and can lead to a safety-critical situation. 
However, previous methods, such as the temporal stereo-based approaches \cite{wang2022dfm, wang2022sts} overlook the importance of moving objects and focus on static scenes.
To confirm that P2D is advantageous in dynamic scenes, we report the detection results of moving objects in Table \ref{table:moving_object}. 
In the table, only objects with a ground-truth velocity higher than 1 m/s are evaluated.
In both BEVDepth and BEVStereo baselines, P2D achieves better performance, especially on the translation (mATE) and velocity error (mAVE). 
We think that this improvement comes from the prediction scheme in P2D, which explicitly provide motion information by forecasting the location of objects.

%==========================================================================================================================================
% ablation-backbone supervision
%==========================================================================================================================================
\begin{table}[t]
\begin{center}
\resizebox{0.48
\textwidth}{!}
{
\setlength{\tabcolsep}{5pt}
\renewcommand{\arraystretch}{1.3}
\begin{tabular}{c|cc|ccccc}
\Xhline{2\arrayrulewidth}

\begin{tabular}[c]{@{}l@{}}\textbf{Prediction} \\ \textbf{supervision}\end{tabular} & \textbf{mAP} $\uparrow$& \textbf{NDS} $\uparrow$& \textbf{mATE} $\downarrow$& \textbf{mAOE} $\downarrow$& \textbf{mAVE} $\downarrow$\\ 
\hline
\hline
            & 0.351 & 0.457 & 0.672 & 0.557 & 0.464 \\
 \checkmark & \textbf{0.360} &\textbf{ 0.474} & \textbf{0.643} & \textbf{0.512} & \textbf{0.412} \\

\Xhline{2\arrayrulewidth}
\end{tabular}
}
\end{center}
\vspace{-5pt}

\caption{Ablation of backbone supervision on prediction loss. The prediction loss does not affect BEV backbone in the model without prediction supervision.} 
\label{table:supervision}

\end{table}
%==========================================================================================================================================

%==========================================================================================================================================
%Ablation - lambda
%==========================================================================================================================================

\begin{table}[t]
\begin{center}
\resizebox{0.47\textwidth}{!}
{
\setlength{\tabcolsep}{2pt}
\renewcommand{\arraystretch}{1.3}
\begin{tabular}{c|cc|ccccc}
\Xhline{2\arrayrulewidth}

\textbf{$\lambda_(p)$} & \textbf{mAP} $\uparrow$ &\textbf{NDS} $\uparrow$& \textbf{mATE} $\downarrow$& \textbf{mASE} $\downarrow$& \textbf{mAOE} $\downarrow$& \textbf{mAVE} $\downarrow$& \textbf{mAAE} $\downarrow$\\ 
\hline
\hline
0.1 & 0.352 & 0.464 & 0.666 & 0.279 & 0.547 & 0.460 & 0.212\\
0.3 & 0.357 & 0.463 & 0.665 & 0.279 & 0.545 & 0.431 & \textbf{0.201}\\
0.5 & \textbf{0.360} & \textbf{0.474} & \textbf{0.643} & \textbf{0.271} & \textbf{0.512} & \textbf{0.412} & 0.217\\

\Xhline{2\arrayrulewidth}
\end{tabular}
}
\end{center}

\vspace{-5pt}
\caption{Ablation of loss balancing weight.} 
\label{table:ablation-lambda}

\end{table}

%==========================================================================================================================================

%==========================================================================================================================================
%Inference
%==========================================================================================================================================

\begin{table}[t]
\begin{center}
\resizebox{0.40\textwidth}{!}
{
\setlength{\tabcolsep}{5pt}
\renewcommand{\arraystretch}{1.3}
\begin{tabular}{c|cc|cc}
\Xhline{2\arrayrulewidth}

\textbf{Methods} & \textbf{mAP} $\uparrow$ &\textbf{NDS} $\uparrow$& \textbf{FPS} $\uparrow$& \textbf{Memory} $\downarrow$\\ 
\hline
\hline
BEVDepth & 0.334 & 0.448 & 8.82 & \textbf{4.26G} \\
P2D & \textbf{0.360} & \textbf{0.474} & \textbf{10.81} & 4.40G \\

\Xhline{2\arrayrulewidth}

\end{tabular}
}
\end{center}

\vspace{-5pt}
\caption{Comparison of inference time and memory usage.} 
\label{table:ablation-time}

\end{table}

\begin{figure*}[t]
\centering
\includegraphics[width=\textwidth]{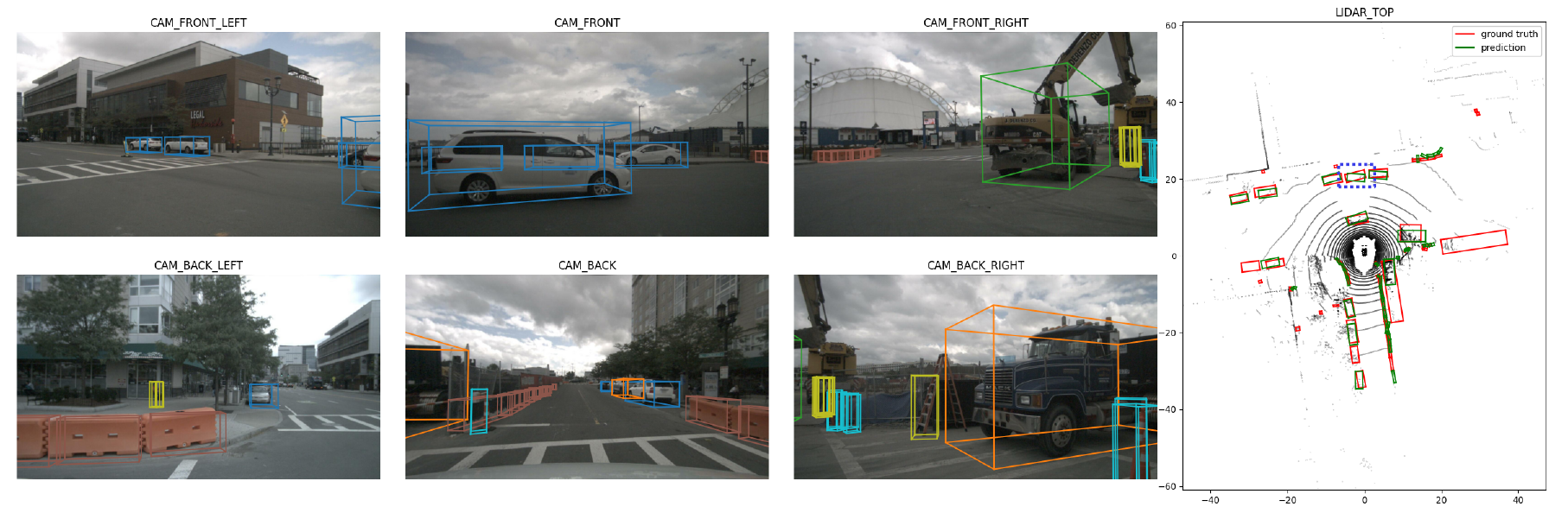}

\caption{
Qualitative results of P2D. The blue dotted rectangle in the BEV view designates the highly occluded object in the image view. Since P2D leverages temporal frames, such an occluded object that appears in the previous frames can be detected.
} 
\label{figure:qualitative}
\end{figure*}

\subsection{Ablation Studies}

We conduct ablation studies to verify the effectiveness of each module and the performance of different hyperparameters. 
We use the nuScenes \textit{val} set and Table \ref{table:ablation} to \ref{table:ablation-time} describe the results of ablation studies.

\noindent \textbf{Prediction head.}\hspace{0.2cm}
Table \ref{table:ablation} demonstrates the ablation of each module in P2D. 
The model with only the prediction head concatenates the prediction results and temporal BEV features without an aggregation strategy.
Adding Prediction Head to the baseline improves mAP by 1.7 \% and NDS by 1.8 points, demonstrating the prediction scheme improves the detection performance. 
Especially the velocity estimation significantly improves by 11.7\%, showing that the prediction strategy helps the model to estimate the motion of objects in the current frame. 

\noindent \textbf{Prediction-guided feature aggregation.}\hspace{0.2cm}
Adding a deformable attention-based feature aggregation also improves mAP by 1.9\% and NDS by 1.1 points, even solely adopted without a prediction head.
We hypothesize that our feature aggregation method merges the features of an object along different timesteps, and thus it is beneficial to make features useful.
Finally, by combining these two modules, our P2D improves mAP and NDS by 2.6 points, showing the effectiveness of our method.

\noindent \textbf{Number of previous frame.}\hspace{0.2cm}
For the fair comparison, we set the number of previous frames as the same and evaluated on the nuScenes \textit{val} set.
As reported in Table \ref{table:previ_frames}, there is still a performance gap between P2D and the baseline BEVDepth with two previous frames. 
Although there is a significant improvement when a previous frame is used due to the benefit from the multi-frame input, adding another previous frame brings only a marginal improvement, demonstrating that increasing previous frames in a naive manner is barely beneficial.
Note that P2D needs at least two previous frames to estimate motion from past frames.

\noindent \textbf{Backbone supervision.}\hspace{0.2cm}
P2D has a prediction loss term that provides supervision for the targets in the current frame to the previous frames. 
We hypothesize that this supervision can guide the backbone to learn how to extract motion-related features from input images.
To verify this, we trained P2D with and without the gradient of the prediction loss in the BEV backbone, and Table \ref{table:supervision} shows the results.
We confirm that with the gradient of the prediction loss on the BEV backbone, the performance improves by 0.9\% and 1.7 points in mAP and NDS, respectively, demonstrating the prediction loss makes the BEV backbone learn motion features.
% the effect of prediction supervision on the BEV backbone and proves our hypothesis. 

\noindent \textbf{Loss balancing weight.}\hspace{0.2cm}
We compare the performance of different values of the loss balancing weight $\lambda_{p}$ in Eq. \ref{eq:loss1}.
As shown in Table \ref{table:ablation-lambda}, the mAP and velocity estimation improves as the value of $\lambda_{p}$ gets larger, proving that the prediction loss helps the model learn motion cues and is beneficial in 3D object detection.

\noindent \textbf{Inference time and memory usage.}\hspace{0.2cm}
Table \ref{table:ablation-time} shows the FPS and GPU memory usage during the inference.
For a fair comparison, both the baseline and P2D use two previous images with the same backbone and image size.
We find that although there is a slight increase in memory usage, P2D runs faster than the baseline by increasing FPS from 8.82 to 10.81.

\subsection{Qualtitative Results}
We visualize a sample case for a qualitative evaluation.
P2D is capable of detecting highly occluded objects, as demonstrated by the object enclosed in the blue dotted box in the top view of Fig. \ref{figure:qualitative}. 
Despite being highly occluded in the current frame, this object has been captured in previous frames, enabling P2D to detect it.
Additional quantitative results are illustrated in Appendix.

\section{Conclusion}
In this work, we propose a novel camera-based 3D object detection using temporal images, namely P2D.
P2D integrates \textit{Prediction} and \textit{Detection} in a single framework to fully benefit from the sequential images. 
P2D improves detection performance, including velocity estimation, and we verified that the motion features obtained by prediction are crucial for 3D object detection.

\noindent{\textbf{Broader impacts.}}\hspace{0.2cm}
P2D has shown that utilizing prediction techniques with motion cues can lead to a significant improvement in the performance of 3D object detection models.
We believe that further research in this area is warranted to explore how best to leverage prediction strategies for more effective 3D object detection. 
In addition, P2D has the potential to inspire the development of 3D object tracking models that place a greater emphasis on motion cues and their role in object detection and tracking.

\section*{Acknowledgements} This work was supported in part by the Korea Agency for Infrastructure Technology Advancement (KAIA) funded by the Ministry of Land, Infrastructure and Transport and the National Research Foundation of Korea(NRF) funded by Korea Government (Ministry of Science and ICT) under Grants RS-2021-KA162184 and 2022R1A2C200494412.

%-------------------------------------------------------------------------

{\small
\bibliographystyle{ieee_fullname}
\bibliography{egbib}
}

% \section{Appendix}
% \input{sections/06_Appendix.tex}

\end{document}